\documentclass{article}
\usepackage{amsmath}
\usepackage{graphicx}
\usepackage{wrapfig}
\usepackage{subcaption}
\usepackage{algorithm}
\usepackage{algorithmic}

\usepackage[main, final]{neurips_2026}

\usepackage[utf8]{inputenc} 
\usepackage[T1]{fontenc}    
\usepackage{hyperref}       
\usepackage{url}            
\usepackage{booktabs}       
\usepackage{amsfonts}       
\usepackage{nicefrac}       
\usepackage{microtype}      
\usepackage{xcolor}         
\usepackage{makecell}

\newcommand{\name}{Ariadne}

\title{Trajectory Planning without Trajectory Data: A Manifold-Guided Approach}

\author{
\begin{tabular}{c}
\textbf{Silong Yong}$^{1}$ \quad \textbf{Anji Liu}$^{2}$ \quad \textbf{Cunxi Dai}$^{1}$ \quad \textbf{Carl Busart}$^{3}$ \\
\textbf{Guanya Shi}$^{1}$ \quad \textbf{Yilun Du}$^{4}$ \quad \textbf{Katia Sycara}$^{1}$ \quad \textbf{Yaqi Xie}$^{1}$ \\
\\
\normalfont
$^{1}$Carnegie Mellon University \quad
$^{2}$National University of Singapore \\
\normalfont
$^{3}$DEVCOM Army Research Laboratory \quad
$^{4}$Harvard University
\end{tabular}
}

\begin{document}

\maketitle

\begin{abstract}

  A common way for trajectory planning is to leverage generative models trained on large collections of expert trajectories. At inference time, the model generates executable trajectories by conditioning on task goal constraints. 
  However, trajectory-based methods rely on costly supervision, scale poorly with sequence length, and often generalize poorly to unseen constraints such as novel start-goal pairs. We propose an alternative to learn the underlying state-space manifold and use the geometry of the manifold for trajectory planning. This approach requires only state observations and enables generalization to unseen constraints by constructing trajectories on the learned manifold of the state space.
Experiments on Maze2D and robotic motion-planning benchmarks show that \textbf{\name}
constructs feasible paths from state-only supervision and generalizes to unseen
start-goal combinations. On high-dimensional dual-arm planning, it remains
competitive with trajectory-supervised and classical planners, while requiring no
trajectory data for training. Code can be found in \url{https://github.com/SilongYong/Ariadne/}.
\end{abstract}


\section{Introduction}
\label{sec:intro}

\looseness=-1
Trajectory planning aims to find a time-parameterized sequence of states that satisfies environment dynamics, task objectives, and constraints. Classical approaches typically rely on search or optimization over explicit dynamics models~\cite{tassa2012synthesis, posa2014direct, kavraki2002probabilistic, kuffner2000rrt}, which can become computationally expensive in high-dimensional spaces.

\begin{wrapfigure}[21]{r}{0.5\textwidth}
  \centering
    \includegraphics[width=\linewidth]{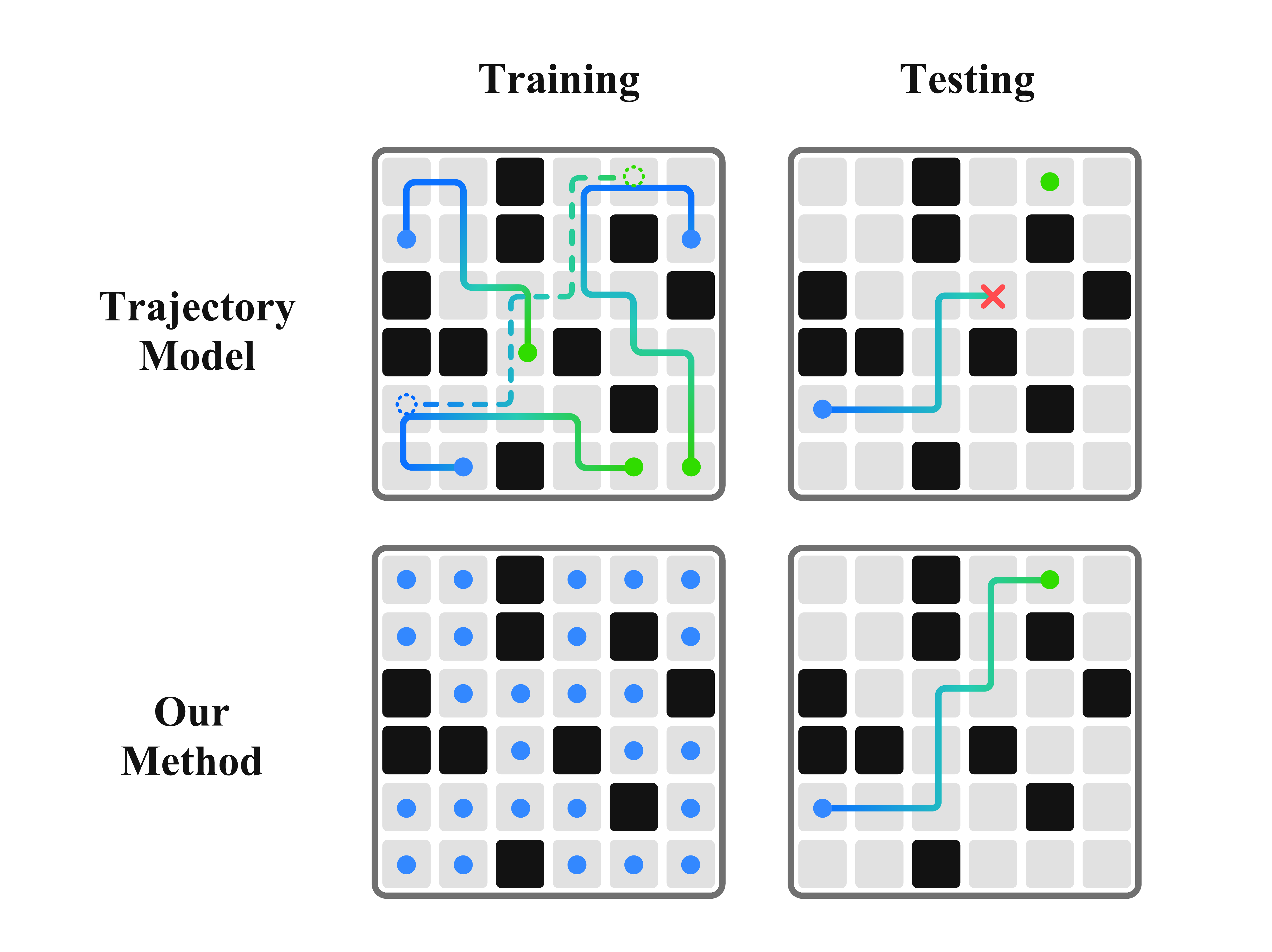}
    \caption{
      Trajectory-based model can't generalize towards unseen combination of start and goal state even if both states are presented in the training data. In contrast, our method is able to generalize under such condition without seeing any trajectory data. 
    }
    \label{fig:teaser}
\end{wrapfigure}

\looseness=-1
Recent learning-based methods instead leverage generative models, such as diffusion models~\cite{ho2020denoising}, to learn trajectory distributions from offline data, enabling flexible plan synthesis via conditional sampling~\cite{janner2022planning, chi2025diffusion, chen2024diffusion, luo2025generative}. By bridging sampling and planning, these methods model distributions over trajectory sequences. However, the trajectory space is exponentially large and highly structured, making it impractical to achieve sufficient coverage with finite data. As the sequence length increases, the number of feasible trajectories grows combinatorially, requiring prohibitively large datasets to learn a reliable model (Fig.~\ref{fig:teaser}, top-left). Empirically, this leads to poor generalization when conditioning on unseen start–goal pairs: the generated trajectories often fail to reach the goal or violate environment constraints, even when the constituent states have been observed during training (Fig.~\ref{fig:teaser}, top-right).

\looseness=-1
We aim to improve the generalizability of generative-model-based trajectory planning while reducing the need for extensive trajectory supervision. Instead of learning from expert trajectories, we consider a setting where only state observations are available, i.e., a generative model is trained to capture the distribution of valid states (Fig.~\ref{fig:teaser}, bottom row). Under the manifold hypothesis~\cite{tenenbaum2000global, hinton2006reducing, he2023manifold, chung2022improving}, these states are assumed to lie on a low-dimensional manifold embedded in the ambient space, i.e., a space that can be curved or complex overall, but looks locally like a flat Euclidean space. While learning such state distributions is well understood, a key challenge remains: how can we leverage this learned representation to construct feasible trajectories? We address this by interpreting trajectory planning as searching for valid paths on the learned state manifold, and propose a method that constructs trajectories by explicitly exploiting its geometry.

\looseness=-1
We view trajectory construction as following a vector field defined over the state space, where a trajectory corresponds to an integral curve of this field. Intuitively, at each state, the vector field specifies a local direction of motion, and integrating these directions produces a path connecting the start and goal states. This vector field is constrained to lie on the underlying state manifold, ensuring that the resulting trajectory remains in feasible regions of the state space. This perspective raises three key challenges: identifying the local geometry of the manifold, defining a vector field that respects this geometry, and ensuring meaningful notions of distance for projection.

\looseness=-1
Our key insight is that generative models provide rich geometric information about the data manifold. In particular, the score function captures local first-order structure, while higher-order information reveals the full normal space. Leveraging this, we construct a geometry-aware vector field that remains on the manifold and connects given endpoints. To enable stable projection, we further derive a Riemannian metric from the score function~\cite{bethune2025follow, du2023reduce}. 

\looseness=-1
We first verify the idea in the classical planning domain where a 2D maze is considered. We demonstrate that with only state information provided during training a generative model, our method is able to extract the geometry and perform sequence modeling by providing feasible planned path towards the goal. Our method therefore overcomes a key limitation of learning-based sequence modeling, where models only perform well on start-goal pairs whose trajectories were seen during training.
We further demonstrate that our method can also be adopted to higher-dimensional space where a 6 DoF robot arm is tasked to avoid obstacles in a table top settings. Our method is also extended to two 7 DoF robot arms control problem where the two arms need not only to avoid obstacles but also avoid each other and achieves comparable results to the State-of-the-Art trajectory-learning-based planning method.
Overall, our contributions are as follows:

\begin{itemize}
    \item We extend the manifold hypothesis on score-accessible generative models to not only confirm the existence but also extract general geometry without strong assumptions from it.
    \item We propose a novel trajectory planning framework which requires only state-space training data by utilizing the geometry and introducing an endpoint constraint by connecting the non-linear interpolation formulation with iterative projection method.
   \item We evaluate \textbf{\name} against classical search, trajectory-supervised diffusion
planners, and geometry-based trajectory optimization, and show competitive planning
performance from state-only supervision across Maze2D and robotic motion-planning tasks.
\end{itemize}

\section{Background and Motivation}
\label{sec:bg_motiv}

\looseness=-1
Learning-based trajectory planning is typically formulated as learning from trajectory datasets. Let a \emph{state} be denoted by $x \in \mathcal{X}$, where $\mathcal{X}$ is the state space. A \emph{trajectory} is a time-indexed sequence $\tau = (x_0, x_1, \dots, x_T)$ describing a feasible transition between a start state $x_0$ and a goal state $x_T$. A \emph{task} specifies trajectory-level conditions, such as initial states, goal states, or intermediate constraints. A dataset $\mathcal{D}_{\text{traj}} = \{\tau_i\}_{i=1}^N$ consists of sampled trajectories collected from demonstrations or rollouts. During training, a model learns to approximate the distribution over trajectories $p(\tau)$ from $\mathcal{D}_{\text{traj}}$. At inference time, the model is conditioned on task specifications and generates a trajectory that is evaluated by task completion and feasibility. The most prominent requirements~\cite{janner2022planning} for task specifications are that the trajectory terminates at a specified goal state. This start-goal formulation is fundamental in robotics and motion planning and our work specifically focus on this setting.

As illustrated in Fig.~\ref{fig:teaser} (top-left), trajectory datasets inevitably provide incomplete coverage of the space of feasible trajectories. Certain valid trajectories, although physically realizable, may be absent from the dataset (shown as dashed curves), due to limited data collection or biased sampling.

This incomplete coverage leads to a well-recognized generalization issue in generative planning models~\cite{janner2022planning, chen2024simplehierarchicalplanningdiffusion, liang2023adaptdiffuserdiffusionmodelsadaptive, chen2025extendableplanningmultiscalediffusion}. Because the model learns a trajectory-level distribution, it becomes biased toward motion patterns observed during training, i.e., biased patterns in the training set. At inference time, when tasked with generating trajectories that were not explicitly present in $\mathcal{D}_{\text{traj}}$, the model often fails to produce valid solutions (Fig.~\ref{fig:teaser}, top-right). This failure arises from trajectory-level distribution mismatch: the required trajectory lies outside the support of the learned trajectory distribution, even though it may be perfectly feasible in the underlying environment.

We propose a new paradigm to address this limitation. We argue that many trajectory-level out-of-distribution (OOD) failures originate from bias in trajectory data rather than from insufficient state coverage. While trajectory samples may be sparse or incomplete, the underlying state distribution often exhibits significantly broader coverage (Fig.~\ref{fig:teaser}, bottom-left). That is, even if certain trajectories are missing, their constituent states may still lie within the observed state space. This observation motivates a fundamental question: \textbf{can we learn planning from state data alone, without relying on trajectory supervision?}

\looseness=-1
Eliminating trajectory supervision immediately raises a key challenge: without access to explicit path information, how can trajectories be reconstructed at inference time? We observe that, for most planning problems, valid trajectories correspond to smooth curves in the state space that remain within dynamically feasible, low-energy regions. If a model can learn the geometric and energetic structure of the state space from state-only data, then trajectory construction can be framed as a curve generation problem under task constraints (e.g., specified endpoints). Moreover, collecting state data is often substantially easier and less restrictive than collecting full trajectory demonstrations, making this paradigm more scalable in practice.

Formally, given a learned state-space model and task-level conditions such as start and goal states, our objective becomes: how can we construct a feasible curve at inference time that obeys these conditions while remaining in valid regions of the state space? We present \textbf{\name}\footnote{Ariadne refers to Ariadne's thread in Greek mythology, which guides Theseus through the labyrinth. We use it to reflect our goal of constructing a path through complex state spaces.} \ below, which exploits the geometry of the learned data manifold and formulates trajectory construction as integrating an ordinary differential equation (ODE) that evolves along it.

\begin{figure*}[ht]
  \vskip 0.2in
  \begin{center}
    \includegraphics[width=\linewidth]{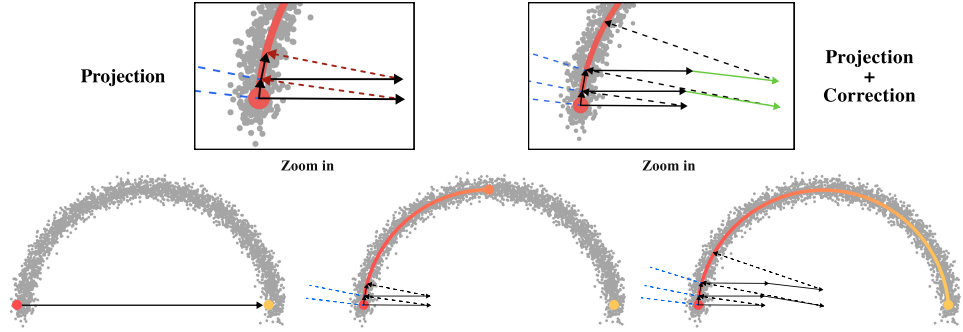}
    \caption{
    \looseness=-1
      Method overview. We take Uniform Circular Gaussian (UCG) as an example. We assume we have a diffusion model trained on such data and aim to use it for constructing a sequence of states that goes from the red dot to the yellow dot. \textbf{Left:} straight line interpolation between the two points doesn't consider the information of the manifold, but enforce the endpoint constraint that the line would start from the red dot and end at the yellow dot. \textbf{Middle:} We can iteratively move along the straight line and use the geometry stored in the diffusion model to project the movement back towards the manifold. However, the projection would shrink the actual moving distance towards the goal, therefore resulting in satisfying manifold constraint but overlooking the endpoint constraint. \textbf{Right:} In order to satisfy both the manifold constraint and the endpoint constraint, we add a correction term derived from an ODE perspective and successfully construct a path represented by the ODE that satisfies both the constraints. 
    }
    \label{fig:main}
  \end{center}
\end{figure*}

\section{Trajectory Construction on the Learned Manifold}
\looseness=-1
Before proceeding, we briefly clarify the notion of the data manifold.
Let $\mathcal{X} \subset \mathbb{R}^d$ denote the ambient state space.
Given state-only training data sampled from valid system behaviors,
we assume that feasible states concentrate around a lower-dimensional
subset $\mathcal{M} \subset \mathcal{X}$, which we refer to as the
\emph{data manifold}.
Intuitively, $\mathcal{M}$ represents the set of dynamically feasible
and physically valid states of the system.

Rather than assuming explicit access to $\mathcal{M}$,
we rely on a learned generative model to implicitly characterize its local geometry.
In particular, the local geometry, i.e. normal and tangent spaces at a point, are defined with respect to this learned geometric structure.


We first consider the problem of constructing a trajectory that remains on the data manifold $\mathcal{M}$, ignoring endpoint constraints for now. A natural approach is to iteratively follow a reference direction while projecting the update back onto the local tangent space of the manifold.

\looseness=-1
Let $x_0 \in \mathcal{M}$ be the initial state and $x_1$ be the desired endpoint.
We define a constant direction vector $x_1-x_0$.
If no manifold constraint were imposed, integrating this direction
would produce the straight-line path
\[
x(t) = x_0 + t (x_1 - x_0),
\]
which exactly reaches the endpoint.

However, this straight-line path will in general leave the manifold.
To enforce the manifold constraint, we adopt a local move-then-project strategy.
At each step, we first move along the constant direction:
\begin{equation}
x_t' = x_{t-1} + (x_1 - x_0)\Delta t.
\label{eq:const}
\end{equation}
The intermediate point $x_t'$ may lie off the manifold.
We then remove the normal component w.r.t the manifold of the displacement to project the motion back onto the tangent space at $x_{t-1}$.

\looseness=-1
Let $\mathbf{V}$ denote an orthonormal basis spanning
the normal space at $x_{t-1}$.
The projected update is given by

\begin{equation}
\begin{aligned}
x_t
&=
x_t'
+
\sum_{n\in\mathbf{V}}
\big[
-\langle x_t' - x_{t-1}, n\rangle n
\big],
\end{aligned}
\label{eq:iter_proj}
\end{equation}

which subtracts the normal components of the displacement.
Equivalently, this corresponds to projecting the update
onto the tangent space $T_{x_{t-1}}\mathcal{M}$.

By repeatedly applying Eq.~\eqref{eq:const} and Eq.~\eqref{eq:iter_proj},
the trajectory remains locally tangent to the manifold at every step.
Intuitively, each iteration removes the component that would push the state
away from $\mathcal{M}$.
As illustrated in Fig.~\ref{fig:main} (middle), this generates a smooth curve
that stays on the manifold while being guided by the global direction
from $x_0$ to $x_1$. While this construction encourages the trajectory to remain on the manifold, it does not strictly guarantee it. In particular, if the reference direction aligns with the normal space, the update may push the state away from the manifold, and the projection step may not fully compensate for this effect. As a result, deviations from the manifold can accumulate over time. Nevertheless, we find that this formulation works well in practice in most cases.

More importantly, even when the trajectory approximately stays on the manifold, this construction does not guarantee satisfaction of task-level endpoint constraints. The projection step can alter the global direction of motion, causing the trajectory to deviate from the intended target and fail to reach the desired endpoint. This reveals a key limitation: local geometric corrections alone are insufficient to control the global behavior of the trajectory. In the next section, we address this issue by explicitly incorporating endpoint constraints into the trajectory construction.


\section{Incorporating Endpoint Constraints}

\looseness=-1
The original move-then-project procedure uses endpoint information through the direction vector pointing toward the target. However, once this direction is projected onto the tangent space, its magnitude and orientation are altered. In particular, the projection removes the normal component of motion, but this removal is not controlled with respect to the remaining progress toward the goal. As a result, although each infinitesimal update locally moves the state “toward” the endpoint in a geometric sense, the accumulated effect may significantly distort the global trajectory. In extreme cases, the projected dynamics may stall or converge to a point that satisfies the manifold constraint but fails to reach the specified endpoint.


\begin{wrapfigure}[24]{r}{0.5\textwidth}
\centering
    \includegraphics[width=\linewidth]{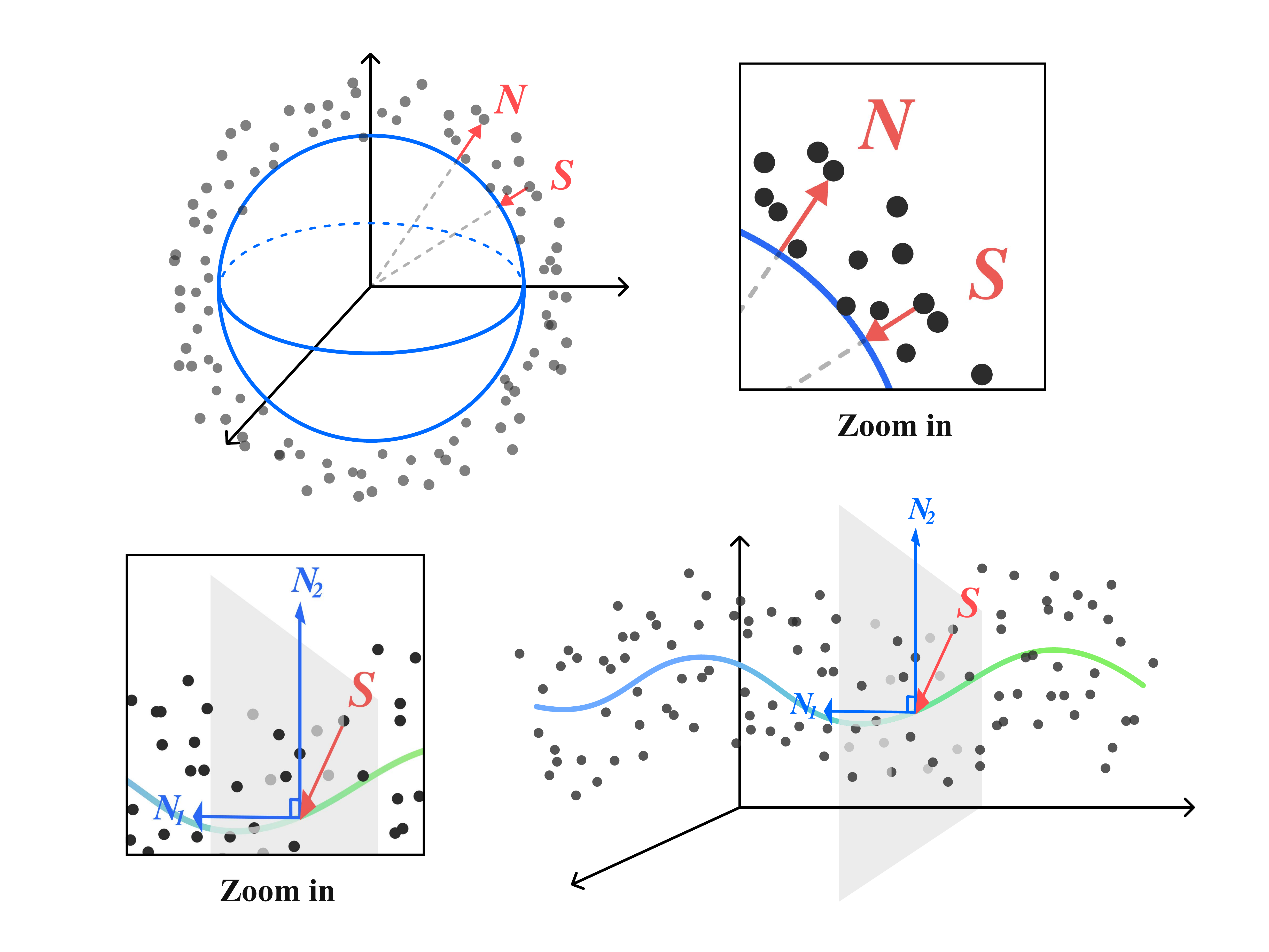}
    \caption{
    \looseness=-1
      While the score function provided by the diffusion model points toward the data manifold and can be viewed as an approximation of the normal direction of a (D-1)-dim manifold embedded in D-dimensional space, it no longer holds when we're facing a more general and realistic case where the manifold might be even lower dimensional. We therefore propose to use the Jacobian of the score function for normal estimation as shown on the right.
    }
    \label{fig:score}
\end{wrapfigure}

\looseness=-1
\paragraph{Roadmap.}
In its discrete iterative form, the move-then-project procedure enforces the manifold constraint but does not guarantee satisfaction of the endpoint condition. To address this issue, we first reinterpret the move-then-project algorithm as the discretization of a specific ordinary differential equation (ODE). In particular, we show that the iterative projection update corresponds to an operator-splitting scheme for a certain projected flow.

\looseness=-1
Building on this formulation, we then analytically modify the induced ODE so that its integral curves both (i) remain on the manifold and (ii) satisfy the prescribed start and endpoint constraints. In this way, endpoint satisfaction is no longer an emergent property of an iterative projection algorithm, but a structural property of the designed dynamics.

\looseness=-1
We'll briefly introduce the operator-splitting methods and then discuss the specific ODE to which the move-and-project procedure corresponds.

\subsection{Operator Splitting}

Operator splitting provides a convenient way to decompose a dynamical system into
multiple sub-dynamics that can be evolved sequentially. For an ODE
\begin{equation}
\frac{dx(t)}{dt}=A(x(t))+B(x(t)),
\label{eq:operator_split}
\end{equation}
a first-order splitting scheme alternates between the dynamics induced by $A$ and
$B$ over each integration step~\cite{Trotter1959OnTP, strang1968construction, macnamara2017operator}.

In our formulation, operator splitting is used primarily as a conceptual tool to
separate two naturally distinct operations in trajectory construction: reference
motion and manifold projection. Specifically, the reference motion corresponds to
$A$, while the projection onto the learned local manifold geometry corresponds to
$B$. This decomposition allows us to reinterpret the iterative move-then-project
procedure as the discretization of an ODE, and subsequently introduce the endpoint
correction in a structured way. The explicit Euler realization and full derivation
are provided in Appendix~\ref{app:ode_derivation}.


\subsection{Interpreting Iterative Projection as ODE}
\looseness=-1
Following the previous discussion, we give the ODE view of the move-then-project procedure in Eq.~\eqref{eq:const} and Eq.~\eqref{eq:iter_proj}.
\begin{equation}
    \frac{dx}{dt} = x_1 - x_0 + 2t(1-t)\frac{d}{dt}\varphi(x_t),
    \label{eq:proj_only}
\end{equation}
where
\begin{equation}
    \varphi(x_t) = \sum_i
    \sum_{n\in\mathbf{N_i}}\frac{(-nn^T)dx}{2t(1-t)\epsilon}\Delta t.
    \label{eq:phi}
\end{equation}
Here $i$ denotes the discrete integration step when iterating the ODE to $x_t$, $\epsilon$ is a factor that controls the scale of projection (also cancels out the $
\Delta t$ that is not in the original projection form) and $\mathbf{V}_i$ denotes an orthonormal basis of the normal subspace at the current state $x_t$.
Intuitively, this formulation views move as $A$ and projection as $B$ in Eq.~\eqref{eq:sum_ode}. 
Please find the full derivation in Appendix~\ref{app:math}.

This perspective reveals that endpoint satisfaction is not automatically guaranteed: the induced ODE enforces invariance to the manifold but does not, by itself, enforce endpoint conditions.

To properly incorporate endpoint constraints, we must analytically design a vector field whose integral curves both (i) remain on the manifold and (ii) satisfy the prescribed endpoint conditions. In other words, we need to define a velocity field that contains the RHS terms in Eq.~\eqref{eq:proj_only}, but also enforces the integral of the velocity field to be a curve anchored at the start and end points.

We therefore propose an ODE that obeys the two rules
\begin{equation}
    \frac{dx}{dt} = x_1 - x_0 + (2-4t)\varphi(x_0, x_1, t) + 2t(1-t)\frac{d}{dt}\varphi(x_0, x_1, t).
    \label{eq:cp_ode}
\end{equation}
This new formulation would still satisfy the manifold constraint since we can view $x_1 - x_0 + (2-4t)\varphi(x_0, x_1, t)$ as $A$ and $2t(1-t)\frac{d}{dt}\varphi(x_0, x_1, t)$ as $B$ in the operator-splitting method. As long as there exists a projection step, i.e.  $2t(1-t)\frac{d}{dt}\varphi(x_0, x_1, t)$ or $B$, in our formulation, the constructed path represented by the ODE will always be on the manifold. In the continuous formulation, the projection component is designed to preserve local
manifold adherence while the correction term enforces the endpoint constraint. In
practice, numerical discretization and imperfect geometry estimation can introduce
small residual manifold drift and endpoint error, which we quantify experimentally in Tab.~\ref{tab:yam_ucg}.

Intuitively speaking, the added term $(2-4t)\varphi(x_0, x_1, t)$ gradually corrects/adds back the removed components from the projection when evolving the ODE. The full derivation can be found in the Appendix~\ref{app:math}.
We name Eq.~\ref{eq:cp_ode} a Correction-Projection ODE. As can be seen from Fig.~\ref{fig:main}, the resulting path constructed using this method indeed stays on the semicircle and it starts from the given start point and ends at the given endpoint.

\subsection{Numerically evolving CP-ODE}
Given the above analysis and derivation, we now give the overview of our proposed method.

We use Euler method with operator-splitting to solve the ODE Eq.~\ref{eq:cp_ode}, namely
$$
\begin{aligned}
    x_t' &= x_{t-1} + (x_1-x_0)\Delta t, \\
    x_t''&= x_t' + (2-4t)\varphi(x_t')\Delta t,\\
    x_t &= x_t'' + 2t(1-t)\frac{d}{dt}\varphi(x_t'')\Delta t.\\
\end{aligned}
$$

The pseudo-code can be found in Algo.~\ref{alg:cpode}. In practice, there are three hyperparameters to set to evolve the CP-ODE, namely the initial arc-length (or step size for evolving the ODE in state space), the projection scale $\epsilon$ and the normal selection threshold (how negative an eigenvalue can inform a normal direction). We provide ablation studies on these three factors in Sec.~\ref{subsec:ablation}.

\paragraph{Manifold adherence and endpoint satisfaction.}
We quantitatively verify the two roles of the Correction-Projection dynamics on the
UCG example in Fig.~\ref{fig:main}. Since the ground-truth semicircular manifold is
known, we measure the average Euclidean distance of the generated path to the manifold,
together with the final endpoint error.
As shown in Table~\ref{tab:yam_ucg}, on UCG, the straight-line reference satisfies the endpoint constraint by construction
but deviates substantially from the ground-truth manifold. Projection-Only exhibits
the opposite behavior: it remains closest to the manifold but fails to reach the target.
\textbf{\name} balances the two objectives, substantially reducing manifold deviation relative
to the straight-line reference while reducing endpoint error by an order of magnitude
relative to Projection-Only. This supports the complementary roles of the projection
and correction terms in the proposed dynamics.

\subsection{Manifold Hypothesis and Geometry from Generative Models}

The only missing component in our formulation is the normal directions required for projection. We aim to extract this information from a generative model trained on state-space data. Let $p(x)$ denote a data distribution in $\mathbb{R}^D$, with samples concentrated near a $k$-dimensional manifold $\mathcal{M}$. We assume access to the score function $s(x) = \nabla_x \log p(x)$.

The score captures first-order geometry and typically points toward higher-density regions, providing a useful approximation of normal directions when the manifold has co-dimension one. However, for general manifolds with higher co-dimension, the normal space is multi-dimensional and cannot be characterized by the score alone.
\begin{wrapfigure}[24]{r}{0.5\textwidth}
\centering
    \includegraphics[width=\linewidth]{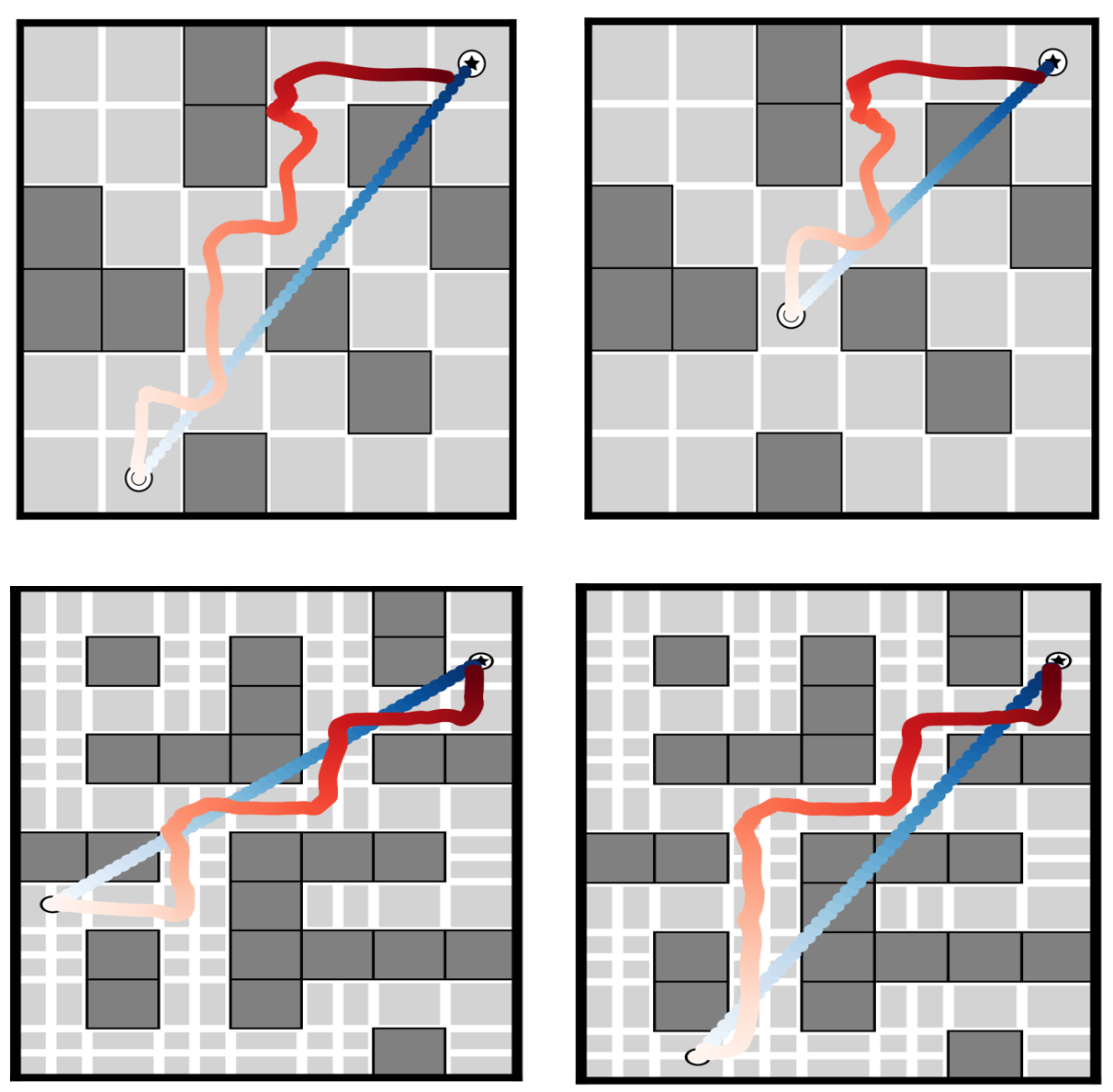}
    \caption{
      Visualization of the planned trajectory on maze. The blue straight line indicates the reference path and the red curve is the output of \textbf{\name}. Light color corresponds to the starting point.
    }
    \label{fig:maze}
\end{wrapfigure}

\looseness=-1
To recover the full normal subspace, we instead consider the Hessian
$$
H(x) := \nabla_x^2 \log p(x),
$$
which encodes second-order geometric information. In particular, under the manifold hypothesis, directions with large negative curvature correspond to the normal space, while near-zero curvature directions correspond to the tangent space. We defer detailed analysis to Appendix~\ref{app:hessian}.



\paragraph{Riemannian metric from the score.}
In addition to identifying normal directions, projection requires a notion of distance. We derive a Riemannian metric from the score function of a pretrained generative model. Under the assumption that the score magnitude is small on the data manifold and increases away from it, we define a position-dependent metric
\begin{equation}
G(x) = \frac{\exp(\|s(x)\|)}{Z} I,
\end{equation}
which remains close to identity on the manifold and grows off it, thereby penalizing deviations. Further details are provided in Appendix~\ref{app:metric}.


\begin{table}[t]
  \caption{
  Maze2D planning results measured by the standard D4RL normalized score.
  Diffuser uses full trajectory supervision, while Diffuser* excludes trajectories
  reaching the evaluation target region. \textbf{\name} is trained only on state observations.
  }
  \label{tab:maze}
  \begin{center}
    \begin{small}
      \begin{sc}
        \begin{tabular}{lcc}
          \toprule
          Method & Medium $\uparrow$ & Large $\uparrow$ \\
          \midrule
          MPPI                  & 10.20  & 5.10   \\
          CQL                   & 5.00   & 12.50  \\
          IQL                   & 34.90  & 58.60  \\
          RRT                   & 45.95  & 10.70  \\
          Follow-the-Energy     & 16.29  & 16.41  \\
          Diffuser*             & 52.70  & 41.98  \\
          Diffuser              & \textbf{120.52} & \textbf{120.37} \\
          \midrule
          \textbf{Ariadne (Ours)} & 55.69 & 50.63 \\
          \bottomrule
        \end{tabular}
      \end{sc}
    \end{small}
  \end{center}
  \vskip -0.1in
\end{table}

\section{Experiments}

\paragraph{Practical considerations.}
In practice, we use a straight-line reference path between endpoints, which we find sufficient for stable performance. We further adopt arc-length-based step sizes to ensure consistent progression along the trajectory, and adapt the implementation for different generative models (e.g., diffusion and flow matching) accordingly. Detailed discussions are provided in Appendix~\ref{app:prac}.

\looseness=-1
\paragraph{Experimental Setup.} We evaluate \textbf{\name} on three planning benchmarks: Maze2D~\cite{janner2022planning, fu2020d4rl}, YAM 6D tabletop manipulation, and DualKuka~\cite{luo2024potential}. In all settings, the model is trained using only state-space data without trajectory or transition supervision. At test time, we generate state sequences connecting start-goal pairs and evaluate performance using standard environment metrics, including task score and collision-free success rate. Detailed dataset construction and evaluation protocols are provided in Appendix~\ref{app:exp_setup}. We compare against classical, trajectory-supervised, and geometry-based planning baselines, including RRT, Diffuser~\cite{janner2022planning}, Follow-the-Energy~\cite{bethune2025follow}, and PBDMP~\cite{luo2024potential}. We provide detailed descriptions of their supervision and inference assumptions in
Appendix~\ref{sec:baseline_details}.

\begin{table}[t]
  \caption{
  \textbf{Left:} YAM 6D motion planning success rate (\%).
  \textbf{Right:} manifold adherence and endpoint accuracy on UCG.
  M-Dist. denotes the average Euclidean distance from the generated path
  to the ground-truth manifold, and E-Err. denotes
  the Euclidean distance between reference endpoint and the resulted endpoint from different methods.
  }
  \label{tab:yam_ucg}
  \begin{center}
    \begin{small}
    \begin{minipage}[t]{0.40\linewidth}
      \centering
      \begin{tabular}{lc}
        \toprule
        Method & SR (\%) $\uparrow$ \\
        \midrule
        Straight Line & 12.31 \\
        \textbf{\name}       & \textbf{60.77} \\
        \bottomrule
      \end{tabular}
    \end{minipage}
    \hfill
    \begin{minipage}[t]{0.56\linewidth}
      \centering
      \begin{tabular}{lcc}
        \toprule
        Method & M-Dist. $\downarrow$ & E-Err. $\downarrow$ \\
        \midrule
        Straight Line   & 0.8800 & 0.0000 \\
        Projection-Only & 0.0322 & 3.3652 \\
        \textbf{\name}         & 0.0798 & 0.3998 \\
        \bottomrule
      \end{tabular}
    \end{minipage}
    \end{small}
  \end{center}
  \vskip -0.1in
\end{table}


\subsection{Experimental Results}
\looseness=-1
\paragraph{Maze2D.} Table~\ref{tab:maze} compares \textbf{\name} against classical, trajectory-supervised,
and geometry-based planners on Maze2D. \textbf{\name} outperforms Diffuser* under the
state-only generalization setting, where the constituent states are observed during
training but trajectories reaching the evaluation targets are withheld. This supports
our hypothesis that broad state-space coverage can enable planning for unseen
start-goal combinations even when the corresponding trajectories are absent from
the training data. When full expert trajectory supervision is provided, Diffuser
achieves substantially higher D4RL normalized scores~\cite{janner2022planning, fu2020d4rl}, highlighting the advantage of
directly observing solutions to the evaluation planning tasks.

\paragraph{Comparison with geometric path optimization.}
Follow-the-Energy~\cite{bethune2025follow} provides the closest geometric comparison to \textbf{\name} because both
methods derive planning information from a learned state-space model. The key distinction
is how this geometry is converted into a trajectory. Follow-the-Energy optimizes an
entire trajectory interpolant under the learned metric, whereas \textbf{\name} constructs the
trajectory online through local Correction-Projection updates.Follow-the-Energy degrades substantially on Maze2D and on two of the three
DualKUKA settings. We observe that trajectory-level interpolation can produce overly
smooth paths in environments with multiple local routes, causing the optimized path
to leave narrow feasible regions. In contrast, \textbf{\name} repeatedly applies local
manifold-guided corrections during path construction. This comparison suggests an
advantage of local Correction-Projection construction over trajectory-level geometric
optimization in these settings.

\looseness=-1
\paragraph{YAM 6D.}
We further evaluate \textbf{\name} on a higher-dimensional tabletop manipulation task.
As shown in Table~2, \textbf{\name} increases collision-free planning success from
12.31\% for the straight-line reference to 60.77\%, demonstrating that the
state-space geometric construction extends beyond the 2D maze setting.
\looseness=-1
\paragraph{DualKUKA.} We next evaluate \textbf{\name} in the higher-dimensional DualKUKA environments, where two
7-DoF robot arms must avoid both workspace obstacles and inter-arm collisions.
Table~\ref{tab:dualkuka} compares against classical planning, trajectory-supervised
diffusion planning, geometric interpolation, and PBDMP. \textbf{\name} achieves the highest
average success rate across the three environments while using only state observations
during training.

Compared with Diffuser, \textbf{\name} obtains substantially higher success rates on all three
settings despite Diffuser receiving full trajectory supervision. One possible explanation
is that the quality of trajectory supervision becomes important in this domain: the expert
trajectories are produced by a classical planner and need not provide short or
optimal solutions, whereas \textbf{\name} does not imitate their trajectory distribution and
instead constructs paths from the distribution of feasible states. \textbf{\name} is also
competitive with PBDMP, although PBDMP additionally uses scene information to support
generalization across environments while our setting learns one planner per scene.

RRT remains a strong classical baseline, but its success rate degrades as the search
problem becomes more difficult and it requires a large number of online collision checks.
A detailed efficiency comparison is provided in Appendix~\ref{sec:efficiency}. 

We additionally evaluate sensitivity to score-model retraining across five independently
trained models and observe consistent performance; full results are provided in
Appendix~\ref{sec:seed_robustness}.

\begin{table}[t]
  \caption{
  Success rate (\%) on DualKUKA across three environments.
  PBDMP takes the scene layout as input, whereas \textbf{\name} learns one planner per scene
  from state-only observations.
  }
  \label{tab:dualkuka}
  \begin{center}
    \begin{small}
      \begin{sc}
        \begin{tabular}{lcccc}
          \toprule
          Method
          & \makecell{$n=7$\\$s=0.14$}
          & \makecell{$n=5$\\$s=0.14$}
          & \makecell{$n=5$\\$s=0.22$}
          & Avg. \\
          \midrule
          Straight Line
          & 23.0 & 14.5 & 26.5 & 21.3 \\
          RRT
          & 47.0 & 36.5 & 50.5 & 44.7 \\
          Diffuser
          & 20.5 & 29.0 & 45.5 & 31.7 \\
          Follow-the-Energy
          & 30.0 & 44.0 & 53.0 & 42.3 \\
          PBDMP*
          & 55.0 & \textbf{50.5} & \textbf{54.0} & 53.2 \\
          \midrule
          \textbf{Ariadne (Ours)}
          & \textbf{57.5} & \textbf{50.5} & 52.0 & \textbf{53.3} \\
          \bottomrule
        \end{tabular}
      \end{sc}
    \end{small}
  \end{center}
  \vskip -0.1in
\end{table}



\subsection{Ablation Studies}
\label{subsec:ablation}
We ablate key hyperparameters of \textbf{\name}, including projection scale, step size, and normal selection threshold as shown in Algo.~\ref{alg:cpode}, on the DualKUKA scenario. 

We ablate the normal-selection threshold, integration step size, and projection scale
in Table~\ref{tab:ablation}. We make the following observations for our method under different hyperparameter settings.
An overly loose normal threshold introduces weak or noisy directions into the estimated
normal space, whereas an overly restrictive threshold produces an incomplete estimate.
Similarly, large integration steps can move the state outside the region where the local
manifold approximation is accurate, while an excessively strong projection can dominate
the reference motion and hinder progress toward the goal. Overall, \textbf{\name} does not rely
on a narrowly tuned hyperparameter configuration. A comprehensive analysis can be found in Appendix~\ref{sec:app_ablation}.

\begin{wrapfigure}[11]{r}{0.5\textwidth}
  \centering
  \includegraphics[width=\linewidth]{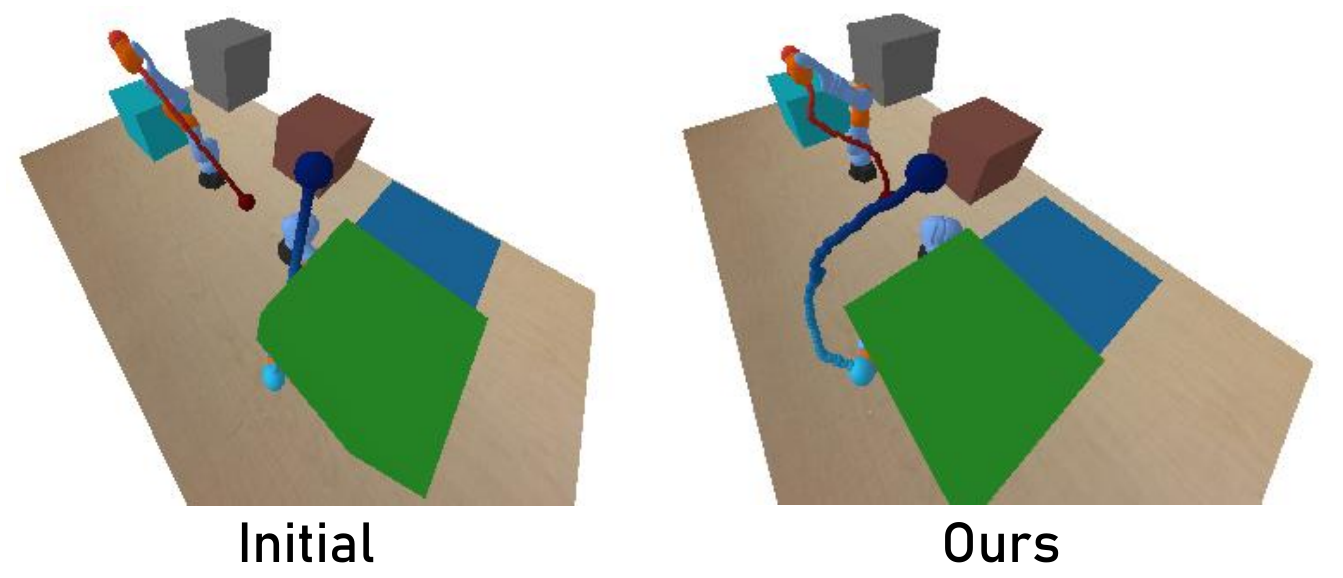}
  \caption{\looseness=-1
  Dualkuka planned trajectory. Dark colors represent start and light colors represent end.}
  \label{fig:dualkuka}
\end{wrapfigure}

\section{Related Work}
For a more detailed review of the related work, please see Appendix~\ref{app:related}.


\looseness=-1
\paragraph{Learning-based Planning.}
Recent advances have introduced sequence modeling techniques into the planning domain~\cite{dalal2024neural, du2021modelbasedplanningenergy}. The key insight is that a trajectory can be viewed as a structured time series, allowing planning to be reformulated as a sequence generation problem. In this paradigm, a model is trained to learn the distribution over trajectories from data, capturing both dynamics and behavioral regularities. At inference time, planning is performed by conditioning the generative model on task-specific information, such as initial states, goals, or constraints, and sampling trajectories that satisfy these conditions.

A central idea behind these approaches is to reinterpret planning as probabilistic inference. Instead of explicitly searching over action sequences, sequence models are trained to approximate the distribution of feasible trajectories. Diffusion-based methods such as Diffuser~\cite{janner2022planning} parameterize this distribution via a denoising process in trajectory space. During inference, planning reduces to conditional sampling, i.e. the model iteratively refines a noisy trajectory while enforcing constraints such as initial states or goal conditions. In this way, trajectory optimization is replaced by guided generation. Subsequent works have improved this framework for long-horizon planning~\cite{luo2025generative} and compositionality~\cite{luo2024potential}, demonstrating the flexibility of generative sequence modeling for planning.



\begin{table}[t]
  \caption{
  Ablation of the three main hyperparameters of \textbf{\name} on DualKUKA
  ($n=7$, $s=0.14$). We vary one parameter at a time while fixing the
  others to their default values. Success rate (\%) is reported.
  }
  \label{tab:ablation}
  \begin{center}
    \begin{small}
      \begin{sc}
        \begin{tabular}{cc|cc|cc}
          \toprule
          \multicolumn{2}{c|}{Normal Threshold}
          & \multicolumn{2}{c|}{Step Size}
          & \multicolumn{2}{c}{Projection Scale} \\
          Value & SR $\uparrow$
          & Value & SR $\uparrow$
          & Value & SR $\uparrow$ \\
          \midrule
          $-12$ & 49.5
          & 0.01 & 54.5
          & 0.4 & 52.0 \\
          $-20$ & \textbf{57.5}
          & 0.02 & \textbf{57.5}
          & 0.5 & \textbf{57.5} \\
          $-25$ & 53.5
          & 0.03 & \textbf{57.5}
          & 0.6 & 54.5 \\
          $-30$ & 51.5
          & 0.04 & 54.0
          & 0.7 & 53.0 \\
          $-35$ & 49.0
          & --   & --
          & 0.9 & 50.0 \\
          \bottomrule
        \end{tabular}
      \end{sc}
    \end{small}
  \end{center}
  \vskip -0.1in
\end{table}

\paragraph{The Manifold Hypothesis.}
The manifold hypothesis~\cite{hinton2006reducing, tenenbaum2000global, lecun2015deep} claims that high dimensional data actually live on a much lower-dimensional manifold embedded in that high-dimensional space, which inspires a series of machine learning techniques~\cite{tenenbaum2000global, hinton2006reducing}. With the development of generative model, it has been proposed to use diffusion models for learning the manifold~\cite{chen2023flow, kapusniak2024metric}, as well as inducing manifold constraints into deep diffusion models~\cite{he2023manifold, chung2024cfg++}. In this work, we extend prior assumptions, i.e. linear subspace, $D-1$-dim manifold, towards a more general setting by using eigendecomposition on the Jacobian of diffusion models for local geometry extraction. Unlike prior works, we do not have a strong assumption over the structure of the data manifold and purely extract the geometry from the learned diffusion model.


\section{Conclusion}

In this work, we introduce \textbf{\name}, a framework for trajectory planning without trajectory supervision. By extracting geometric information from a pretrained state-space diffusion model, \textbf{\name} constructs trajectories under manifold constraints and endpoint constraints. Unlike trajectory-based generative approaches, our method learns only from feasible states and constructs trajectories during inference. Experiments on Maze2D and robotic motion planning benchmarks demonstrate that \textbf{\name} can synthesize feasible trajectories without observing trajectory data during training, while generalizing to unseen start-goal configurations. These results suggest that state-space geometry may provide an alternative inductive bias for trajectory planning beyond explicit trajectory imitation.


\paragraph{Limitations.}
\textbf{\name} assumes that sufficiently nearby states on the learned manifold admit valid local
transitions. State validity alone does not guarantee transition validity in systems with
irreversible, discontinuous, or strongly action-dependent dynamics; extending the model
to state-action or transition tuples is therefore an important direction. Planning quality
also depends on the accuracy of the learned local geometry and can degrade under
insufficient state coverage, multimodal connectivity, or long-horizon problems where a
simple reference path does not capture the correct global topology. In addition, estimating
the normal space requires derivatives of the learned score field, making inference more
computationally expensive as the state dimension grows. Developing more informative reference paths and extending the framework to more complex environments remain important directions for future work.

\begin{ack}
This work has been funded in part by the Army Research Laboratory (ARL) award W911QX-24-F-0049, DARPA award FA8750-23-2-1015, ONR award N00014-23-1-2840, ONR MURI grant N00014-25-1-2116, and the National
University of Singapore under its Start-up Grant (Award No: SUG-251RES2505. The author would like to thank Xinran Zhao for the valuable feedback on the manuscript and Julia Du for the help on visual illustrations.

\end{ack}

\newpage
\bibliographystyle{plain}
\bibliography{reference}



\newpage
\appendix
\section{Extented Related Works}
\label{app:related}
\subsection{Generative Models for Sequence Modeling
}
Generative models have seen remarkable progress in recent years~\cite{rombach2022high, chi2025diffusion, ho2020denoising, achiam2023gpt}. Diffusion models, in particular, have gained significant attention~\cite{janner2022planning, rombach2022high, ho2020denoising, song2019generative, song2020score}. While initially advanced in image domain, works have been done on extending it to sequence modeling including video~\cite{ho2022video, chen2024diffusion}, robotics~\cite{chi2025diffusion} and planning~\cite{janner2022planning, luo2025generative}. In addition, works have also tried to explore the geometry underlying the generative models~\cite{he2023manifold, bethune2025follow, kapusniak2024metric, chen2023flow}. While prior work assumes a learned metric model, we focus on directly extracting the geometry from a trained diffusion model.

\subsection{Latent Interpolation}
Latent interpolation has been widely studied in generative modeling as a tool for exploring learned representations and constructing smooth transitions between data samples~\cite{albergo2025stochastic, slerpkarpathy2022}. 

While the standard way is to do linear interpolation, several advanced methods on latent interpolation~\cite{bethune2025follow, moreau2026probing, kulikov2025flowedit, slerpkarpathy2022, struski2023feature}, considering geodesic paths or spherical interpolation (slerp), have been proposed in diffusion-based models to demonstrate continuity and semantic smoothness of learned latent spaces.

However, latent interpolation primarily operates in the noise space rather than directly enforcing geometric or dynamical feasibility in the ambient data space.
The resulting trajectories are smooth in latent coordinates but are not explicitly constrained to remain connected on a learned data manifold.

In contrast, \textbf{\name} constructs trajectories directly in the data space by explicitly modeling the local geometry of the data manifold and formulating trajectory generation as integrating a constrained ODE.
Rather than interpolating between latent embeddings, we design a dynamical system whose integral curves both remain on the data manifold and satisfy endpoint constraints by construction.

\section{Math formulation for ODE construction}
\label{app:math}
\subsection{Operator Splitting Methods}
\label{app:ode_derivation}
Operator Splitting methods~\cite{Trotter1959OnTP, strang1968construction, macnamara2017operator} are a family of methods that are proposed to solve a specific type of ODE. Consider an ODE whose vector field can be written as the sum of two components:
\begin{equation}
\frac{d x(t)}{dt} = A\!\left(x(t)\right) + B\!\left(x(t)\right),
\label{eq:sum_ode}
\end{equation}
where $A(\cdot)$ and $B(\cdot)$ represent two functions that are easier to handle separately.
Instead of solving Eq.~\eqref{eq:sum_ode} in one shot, Operator splitting suggests that over each time interval of length $\Delta t$, we first evolve the state under $A$
and then evolve it under $B$.

\paragraph{Split subproblems.}
Specifically, for each step we solve the two sub-ODEs sequentially:
\begin{align}
\frac{d x(t)}{dt} &= A\!\left(x(t)\right),
\label{eq:split_A}\\
\frac{d x(t)}{dt} &= B\!\left(x(t)\right).
\label{eq:split_B}
\end{align}

\paragraph{Forward Euler discretization.}
Starting from $x_{t-1}$, we apply a forward Euler step for Eq.~\eqref{eq:split_A}:
\begin{equation}
x_t^{(A)} = x_{t-1} + \Delta t \, A(x_{t-1}),
\label{eq:euler_A}
\end{equation}
followed by a forward Euler step for Eq.~\eqref{eq:split_B} initialized at $x_t^{(A)}$:
\begin{equation}
x_t = x_t^{(A)} + \Delta t \, B\!\left(x_t^{(A)}\right).
\label{eq:euler_B}
\end{equation}
Equivalently, the full update can be written as
\begin{equation}
x_t
=
x_{t-1}
+
\Delta t\, A(x_{t-1})
+
\Delta t\, B\!\left(x_{t-1} + \Delta t\, A(x_{t-1})\right).
\label{eq:split_update}
\end{equation}
\looseness=-1
Operator Splitting works when alternating between sub-dynamics does not introduce uncontrolled curvature or instability relative to the combined flow. We refer the reader to the book about the method for rigorous explanation~\cite{macnamara2017operator}.

\subsection{Iterative Projection}
The iterative projection procedure that we used can be written as follows,
\begin{equation}
    \begin{aligned}
        x_t' &= x_{t-1} + (x_1-x_0)\Delta t, \\
        x_t &= x_t' + \sum_{n\in\mathbf{V}}[-\langle x_t' - x_{t-1}, n\rangle n],
\end{aligned}
\label{eq:iter_proj_app}
\end{equation}
We can rewrite the second line as
\begin{equation}
    x_t = x_t' + \frac{1}{\epsilon}\sum_{n\in\mathbf{V}}[-\langle x_t' - x_{t-1}, n\rangle n]\Delta t
\end{equation}
where $\epsilon$ is a small constant that cancels out the effect of the step size $\Delta t$, i.e., $\epsilon = \Delta t$, $\mathbf{V}$ is the local normal subspace and $n$ are the basis of that subspace. It corresponds to an operator-splitting numerical method for solving the following ODE,
\begin{equation}
    \frac{dx}{dt} = x_1 - x_0 + \sum_{n\in\mathbf{V}}\frac{(-nn^T)dx}{\epsilon},
    \label{eq:ip_ode_app}
\end{equation}
where $\epsilon = O(dt)$, $A(x_t) = x_1 - x_0$ and $B(x_t) = \sum_{n\in\mathbf{V}}\frac{(-nn^T)dx}{\epsilon}$.


\subsection{Correction-Projection ODE}
We start our derivation with a well-established non-linear interpolation formula since it naturally guarantees the start and end points.
\begin{equation}
    x_t = (1-t)x_0 + tx_1 + 2t(1-t)\varphi(x_t),
\end{equation}
where $\varphi(x_t)$ is the non-linear correction term and $(1-t)x_0 + tx_1$ represents a straight-line interpolation. Taking derivative w.r.t t on both sides, we get
\begin{equation}
    \frac{dx}{dt} = x_1 - x_0 + (2-4t)\varphi(x_t) + 2t(1-t)\frac{d}{dt}\varphi(x_t).
    \label{eq:from_interp}
\end{equation}
We now find a correspondence between the iterative-projection ODE and Eq.~\eqref{eq:from_interp}.

Notice that since $\epsilon = O(dt)$, we can draw similarity between the term $\sum_{n\in\mathbf{V}}\frac{(-nn^T)dx}{\epsilon}$ in Eq.~\eqref{eq:ip_ode_app} and $2t(1-t)\frac{d}{dt}\varphi(x_t)$ in Eq.~\eqref{eq:from_interp}. We can set 
\begin{equation}
    \sum_{n\in\mathbf{V}}\frac{(-nn^T)dx}{\epsilon} = 2t(1-t)\frac{d}{dt}\varphi(x_t).
    \label{eq:correspondence}
\end{equation}
Therefore by canceling out $\epsilon$ and $dt$, we have
\begin{equation}
    d\varphi(x_t) = \sum_{n\in \mathbf{V}}\frac{(-nn^T)dx}{2t(1-t)}
\end{equation}
and
\begin{equation}
    \varphi(x_t) = \varphi(x_0) + \int_0^t \frac{d\varphi(x_t)}{ds}ds.
    \label{eq:phi_continuous}
\end{equation}
Since we solve the ODE using Euler method, we can set $\varphi(x_0)  = \mathbf{0}$ and compute the integral in Eq.~\eqref{eq:phi_continuous} as 
\begin{equation}
    \varphi(x_t) = \sum_i
    \sum_{n\in\mathbf{N_i}}\frac{(-nn^T)dx}{2t(1-t)\epsilon}\Delta t,
    \label{eq:phi_compute}
\end{equation}
where $i$ denotes the discrete integration step when iterating the ODE to $x_t$, and $\mathbf{V}_i$ denotes an orthonormal basis of the normal subspace at the current state $x_t$.

\section{Pseudo Code for the proposed Correction-Projection ODE}
Please refer to Algo.~\ref{alg:cpode} for full procedure of evolving our algorithm. Notice that there are three hyperparameters that has to be set for evolving the Correction-Projection ODE, i.e. the initial arc-length (or step size for evolving the ODE in state space), the projection scale and the normal selection threshold. We provide ablation studies on these three factors in Sec.~\ref{subsec:ablation}.

\begin{algorithm}[H]
\caption{Correction Projection ODE}
\label{alg:cpode}
    \begin{algorithmic}
        \STATE {\bfseries Input: } initial arc-length $C$, projection scale $\epsilon$, Normal selection threshold $T_{normal}$, start point $x_0$, end point $x_1$, a flow model learned on state distribution $F$
        \STATE {\bfseries Output: } A feasible plan $P$ with every state in the state space.
    
        \STATE Initialize parameters $t = 0$
        \STATE Initialize state $x \gets x_0$
        \STATE Initialize past state $x_{p} \gets x_0$
        \STATE Initialize post-projection state $x_{pro} \gets x_0$
        \STATE Initialize PHI list $d\varphi = []$
        \STATE Initialize State Path $P = [x_0]$
        
        \WHILE{$t < 1$}
            \IF{$t = 0$}
                \STATE $dt = C$
            \ELSE
                \STATE $dt = \frac{C}{x - x_{p}}$
                \STATE $dt = \min\{dt, 1 - t\}$
            \ENDIF
            \STATE $V_{ref} = x_1 - x_0$
            \STATE Manifold correction scale $M = \min\{\exp(|x - x_0|), \exp(|x - x_1|)\}$
            \STATE $S = F(x,t=0.98)$
            \STATE eigendecomposition: $\frac{\nabla_xS + (\nabla_xS)^T}{2} = Q\Sigma Q^T$
            \STATE $n = \{\frac{Q_i}{M}|\Sigma_i < T_{normal}\}$, $u = sum(\{\Sigma_i |\Sigma_i > 0\})$ 
            \STATE $x_{p} \gets x$
            \FOR{$n_i$ in $n$}
                \STATE $k = -\Sigma_i$ if $-\Sigma_i > -1$ else $k=-\frac{1}{\Sigma_i}$
                \STATE $x = x - \epsilon\frac{u^kM(n_in_i^T)(x-x_{pro})}{||n_i||^2}$ 
                \STATE $d\varphi \gets \epsilon\frac{u^kM(n_in_i^T)(x-x_{pro})}{2t_i(1-t_i)||n_i||^2}$
            \ENDFOR
            \STATE $x_{pro} \gets x$
            \STATE $\varphi = \sum d\varphi$
            \STATE $x = x - 2(1-2t)\varphi dt$
            \STATE $x = x + V_{ref}dt$
            \STATE $t = t +dt$
            \STATE $P\gets x$
        \ENDWHILE
        
        \STATE {\bfseries Return:} $P$
    \end{algorithmic}
\end{algorithm}

\section{Geometry from score function}
\subsection{Hessian and local geometry.}
\label{app:hessian}
We analyze the second-order structure of the log-density via the Hessian
$$
H(x) = \nabla_x^2 \log p(x).
$$
Under the manifold-plus-noise assumption, the log-density is approximately flat along tangent directions of $\mathcal{M}$ and exhibits strong negative curvature along normal directions. Consequently, eigenvectors corresponding to large negative eigenvalues span the normal space, while those with near-zero eigenvalues align with the tangent space.

This decomposition provides a local characterization of the manifold geometry. Deviations from this structure, such as the presence of positive eigenvalues, indicate more complex local structures (e.g., multi-modality or stratified manifolds).

\subsection{Riemannian metric from score function.}
\label{app:metric}
In addition to identifying normal directions of the data manifold, performing projection also requires a notion of distance, i.e., a Riemannian metric defined over the ambient space. We observe that a learned energy function $E_\theta(x)$ from diffusion or score-based models naturally provides such geometric information~\cite{bethune2025follow, du2023reduce}.

Ideally, the learned energy satisfies two properties: (1) $E_\theta(x)$ is small on the data manifold and increases as $x$ moves away from it; and (2) the gradient magnitude $|\nabla_x E_\theta(x)|$ is close to zero when $x$ lies on the manifold. Under these assumptions, the norm of the score function serves as a proxy for the distance to the manifold since 
$s(x) = -\nabla_x E_\theta(x)$.

Motivated by this observation, we define a Riemannian metric
\begin{equation}
G(x) = \frac{\exp\big(|s(x)|\big)}{Z} I,
\label{eq:score_metric}
\end{equation}

where $Z$ is a normalization constant and $I$ is the identity matrix.

By construction,
\begin{equation}
G(x) \approx I \quad \text{when } x \in \mathcal{M},
\end{equation}
and $G(x)$ increases as $x$ moves away from the manifold. This induces a geometry in which off-manifold directions incur larger metric cost, effectively penalizing deviations from $\mathcal{M}$.

Therefore, under the assumption that $|\nabla_x E_\theta(x)| \approx 0$ on the data manifold, a Riemannian metric can be directly derived from a learned generative model without explicit manifold supervision.

\section{Practical Considerations Full Details}
\label{app:prac}
\paragraph{Reference path selection.} Notice that $x_1 - x_0$ is a direct outcome of the linear interpolation between $x_1$ and $x_0$. We may also use different parametrizations for the interpolation such as SLERP~\cite{slerpkarpathy2022}, non-perfect trajectories generated by trajectory-based generative models~\cite{luo2024potential, janner2022planning}, or parametrized curve like
\begin{equation}
    x(t) = (1-t)x_0 + tx_1 + (\sum_{k=1}^{m}\alpha_k \sin(k\pi t))d,
\end{equation}
where $m$ and $\alpha_k$ are hyperparameters deciding the shape of the curve and $d$ is chosen to be a perpendicular direction w.r.t the straight line $(1-t)x_0 + tx_1$.

However, we do find that in practice straight-line reference path is sufficient enough for \textbf{\name} to provide reasonable results and we stick to the straight line for our experiments.

\paragraph{Step size and parameter $t$.} In order to ensure the connectivity in the state space, instead of choosing a fixed time interval $dt$, we should instead do a change of variable and consider a fixed step size $ds$, the time interval should then be $dt = \frac{ds}{||f||}$, where $f$ is the RHS of Eq.~\ref{eq:cp_ode}. 
\paragraph{Flow Matching and Diffusion.} While diffusion models directly predict the score function and thus lead to simple Hession computation, the implementation over a Flow-Matching model should slightly be changed to use $\frac{(J+J^T)}{2}$ as the approximation of the Hession matrix where $J$ is the Jacobian of the output of Flow-Matching w.r.t its input.

\section{Experimental Setup Details}
\label{app:exp_setup}
We test our algorithm in the Maze 2D environment and only train our model using the state-space data of the maze without any trajectory data that inform connectivity. We report the standard score gathered in the environment as the evaluation metric.

We further validate \textbf{\name} on the YAM 6D tabletop setup, where a 6-DoF robotic arm operates in a 3D workspace containing tabletop obstacles. We evaluate the success rate of collision-free motion planning. Similar to Maze 2D, our model is trained using state-only supervision, with no trajectory data and no transition information.

We construct the training set by sampling arm configurations and applying forward kinematics to obtain end-effector and link poses. We then run a collision filter against the tabletop and obstacles, and retain only collision-free states as feasible training data.

For each test instance (start-goal pair), we plan a state sequence using the learned model and execute the resulting plan under the environment’s standard collision and validity checks. We report the task success / score defined by the benchmark (same scoring protocol as the environment), emphasizing that performance is achieved despite training without trajectories.

For all experiments, we use 0.02 as the step size for evolving the ODE. We use -0.0001 as normal threshold and 1.0 as projection scale for Maze 2D, -1.0 as normal threshold and 0.5 as projection scale for YAM 6D and -20.0 as normal threshold and 0.5 as projection scale for DualKUKA 14D. We found that as the dimension of the problem goes up, the absolute value of eigenvalues goes up, and therefore bigger normal threshold. For close manifold normal direction estimation, we use the timestep $0.98$ as input to the flow-matching model that we used as it is trained to estimate a time-dependent vector field and close to 1 could provide a reasonable normal estimation for the real data manifold.


We further adopt DualKuka~\cite{luo2024potential}, a more challenging scenario where two 7-Dof robot arm is tasked to avoid obstacles with more complex layout as well as not colliding with each other. The state space is defined to be all the collision-free end effector pose. Similar to the YAM 6D setting, each test instance is a given start-goal pair for both the arm and we aim to generate a state sequence using our algorithm. The success rate is computed as the \textbf{one-shot} success of a collision-free and goal-reaching path generation. We spawn 7 and 5 obstacles following the standard data generation process of DualKUKA for quantitative evaluation. The half size of the block is set to 0.14, 0.22 respectively for the three scnarios reported in Tab.~\ref{tab:dualkuka}. Qualitative results are given with 5 obstacles with 0.22 as half size as it gives better visualization of how our method provides obstacle avoidance capability, i.e. manifold following, for a path that has collision. The ablation study is done on 7 obstacles with 0.14 as the half size. For our results, we use the end effector space with XYZ position and euler angle for the state representation. We found that as this representation reveals the essential problem dimension, our method is able to provide comparable results to State-of-the-Art trajectory-based methods.

For baselines, since we follow the tradition as Diffuser~\cite{janner2022planning} to train single planner for a single scene, we restrict the evaluation to a single scene and evaluate the baselines also on this restricted scene even if they offer generalizable capabilities to take as input different scenarios' layout.

\section{Training Details}
For training the flow-matching model, we use a MLP with 4 hidden layers and 128 as the latent dimension for Maze 2D and YAM 6D. For DualKUKA, we use MLP with 8 hidden layers and 1024 as hidden dimension. The learning rate for Maze 2D medium, Maze 2D large, YAM 6D and DualKUKA are respectively $1e-4$, $5e-4$, $1e-4$ and $2e-5$. The batch size for Maze 2D and YAM 6D is 64 and for DualKUKA is $1024$. We transform the joint angle representation of DualKUKA into end effector space and represent it by euler angle. The inference of our algorithm is conducted in end effector space and use IK with the joint angle reference of the start point to derive the joint angles for both the KUKA arms. All the training is conducted on a single NVIDIA RTX 6000 Ada Generation GPU and requires from 2 hours to 12 hours depending on the tasks and dataset size. We follow the standard way to generate the Maze data, randomly uniformly sampled for the YAM 6D environment and uniformly sampled for the DualKUKA environment. The GPU memory needed for training such model is only at around 400M.

\section{More Quatitative Results}
\begin{figure}[h]
  \centering
  \includegraphics[width=0.85\linewidth]{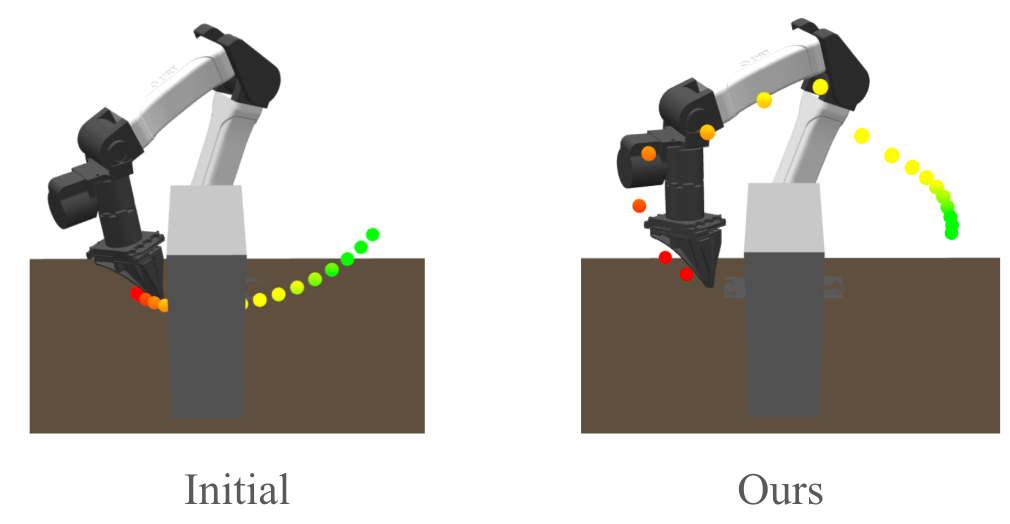}
  \caption{Visualization of the planned trajectory. Color of the way-points specify task progress, where initial state is green and final state is red.}
  \label{fig:yam6d}
\end{figure}

In Fig.~\ref{fig:yam6d}, we visualize a representative planned trajectory produced by \textbf{\name} in the YAM 6D scene: the plotted path shows the end-effector (and/or configuration) sequence staying within the collision-free manifold and routing around obstacles to connect the specified start and goal.

\section{Extented Analysis on Ablation Studies}
\label{sec:app_ablation}
\paragraph{Projection scale.}
We compare multiple scaling factors in Tab.~\ref{tab:ablation} and observe that performance peaks at a moderate scale, while both smaller and larger values lead to degraded results. This can be attributed to a trade-off between following the reference direction and enforcing manifold constraints. When the projection scale is too small, the update is dominated by the reference direction, causing the trajectory to drift away from the manifold and resulting in infeasible states. In contrast, when the projection scale is too large, the accumulated projection corrections dominate the update, significantly altering the direction of motion and causing the trajectory to drift away from the reference path. This hinders effective progress toward the goal. A moderate projection scale balances these effects, leading to more stable and effective trajectory construction.

\paragraph{Step size.}
We evaluate step sizes of 0.01, 0.02, 0.03 and 0.04 in Tab.~\ref{tab:ablation}. We observe that moderate step sizes achieve the best performance, while both smaller and larger values lead to degraded results. This can be attributed to a trade-off in the discretized update dynamics. Smaller step sizes provide a more accurate approximation to the underlying continuous trajectory, but require more iterations to make progress, which can accumulate errors. In contrast, larger step sizes may overshoot the local region where the manifold approximation is valid, moving the state far from the manifold before projection is applied. This weakens the effectiveness of the projection step and degrades performance. Moderate step sizes balance these effects, leading to more stable and effective trajectory construction.

\paragraph{Normal threshold.}
We vary the threshold for selecting normal directions in Tab.~\ref{tab:ablation} and observe that performance peaks at a moderate threshold, while both smaller and larger values lead to degraded results. This can be attributed to a trade-off in estimating the normal subspace. When the threshold is too low, many weak or noisy directions are included as normals, leading to inaccurate projections that distort the trajectory and introduce instability. In contrast, when the threshold is too high, only a small subset of directions is retained, resulting in an incomplete characterization of the normal space. This weakens the constraint and allows the trajectory to drift away from the manifold. A moderate threshold strikes a balance between these effects, yielding a more accurate and stable approximation of the normal space.

\section{Extended Discussion}
\label{app:discussion}
\looseness=-1

A central observation behind \textbf{\name} is that trajectory incompleteness does not necessarily imply state incompleteness. While trajectory datasets often sparsely cover start-goal combinations, feasible states themselves may still be broadly represented. Rather than explicitly learning trajectory distributions, \textbf{\name} constructs trajectories from the geometry of feasible states, enabling planning from state distribution learning.

While our proposed method takes the form of ODE construction, there are several ways that it can be interpreted and connected to existing methods like the transition kernel interpretation of diffusion models~\cite{ho2020denoising}, and transportation between arbitrary distributions~\cite{kulikov2025flowedit, lipman2023flowmatchinggenerativemodeling}.

\textbf{\name} may also be viewed as inducing deterministic local transition dynamics over the learned state manifold. Conceptually related to probability flow ODEs in diffusion models~\cite{ho2020denoising, song2020score}, \textbf{\name} instead considers transport between feasible states while remaining consistent with the learned state distribution.

\paragraph{Transition Kernel.}
The evolution of the ODE that we constructed can also be viewed as a deterministic transition kernel
\begin{equation}
    K(x_{t+dt}|x_t) = \delta(x_{t+dt} - v(x_t, t)dt),
    \label{eq:trans_ker}
\end{equation}
where $v(x_t, t)$ is the RHS of our constructed ODE and $\delta$ is the Dirac delta function.

\looseness=-1
We can view the constructed kernel as a replacement to the Gaussian kernel used in the diffusion process. \textbf{\name} therefore can be viewed as constructing a kernel that we can sample a real data from. Such interpretation can further be extended to stochastic kernels to introduce exploration and multi-modality to the method.

\paragraph{Transportation between Two Distributions.}
\looseness=-1
Similarly, the two endpoint constraints specified in trajectory planning can also be viewed as two Dirac delta function centered at the start and end point respectively. Therefore \textbf{\name} can also be explored together with existing methods like FlowEdit~\cite{kulikov2025flowedit} for high-dimensional sequence generation with only distribution-level constraints.

More broadly, our results raise the question of whether the manifold hypothesis and its geometry may extend beyond trajectory planning to broader sequential domains, which we leave to future work.

\section{Efficiency Comparison}
\label{sec:efficiency}

In addition to planning success, we compare \textbf{\name} with RRT in terms of online
planning efficiency. RRT performs explicit sampling-based search and repeatedly
queries a collision checker to identify feasible paths, whereas \textbf{\name} directly
constructs trajectories from the learned state-space geometry and does not invoke
a collision checker during online path construction. For DualKUKA, $n$ denotes
the number of obstacles and $s$ denotes the obstacle half-size, consistent with
Table~\ref{tab:dualkuka}. We report success rate, the number of collision
checks, and wall-clock planning time in Table~\ref{tab:rrt_efficiency}.

\begin{table}[t]
  \caption{
  Planning efficiency comparison with RRT.
  We report success rate (SR, \%), number of collision checks, and planning time in seconds.
  For DualKUKA, $n$ denotes the number of obstacles and $s$ denotes the obstacle half-size.
  \textbf{\name} does not invoke a collision checker during online trajectory construction.
  }
  \label{tab:rrt_efficiency}
  \begin{center}
    \begin{small}
      \begin{sc}
        \begin{tabular}{llrrr}
          \toprule
          Environment & Method
          & SR $\uparrow$
          & Collision Checks $\downarrow$
          & Time (s) $\downarrow$ \\
          \midrule
          Maze-Medium
          & RRT     & 45.95 & 47,374.67 & 3.08 \\
          & \textbf{\name} & 55.69 & 0         & 0.00145 \\
          \midrule
          Maze-Large
          & RRT     & 10.70 & 90,849.51 & 4.50 \\
          & \textbf{\name} & 50.63 & 0         & 0.00049 \\
          \midrule
          DualKUKA ($n=7,\ s=0.14$)
          & RRT     & 47.0 & 21,039.0 & 8.26 \\
          & \textbf{\name} & 57.5 & 0        & 1.482 \\
          \midrule
          DualKUKA ($n=5,\ s=0.14$)
          & RRT     & 36.5 & 56,311.8 & 25.78 \\
          & \textbf{\name} & 50.5 & 0        & 1.122 \\
          \midrule
          DualKUKA ($n=5,\ s=0.22$)
          & RRT     & 50.5 & 21,862.0 & 7.92 \\
          & \textbf{\name} & 52.0 & 0        & 2.040 \\
          \bottomrule
        \end{tabular}
      \end{sc}
    \end{small}
  \end{center}
  \vskip -0.1in
\end{table}

\section{Detailed Explanation of Baselines}
\label{sec:baseline_details}

We compare \textbf{\name} against planning methods with different supervision and inference
assumptions. RRT is a classical sampling-based planner that explicitly searches the
configuration space and queries a collision checker during planning. Diffuser is a
trajectory-supervised generative planner trained on complete state trajectories.
Diffuser* denotes the restricted variant used in our original Maze2D experiments:
the model observes the evaluation target states during training, but never observes
them as trajectory endpoints, and therefore does not receive trajectory supervision
for the corresponding target-reaching planning problems.

Follow-the-Energy~\cite{bethune2025follow} is the closest geometry-based baseline to \textbf{\name}. It learns an
energy model from state observations, derives a Riemannian metric from the learned
energy landscape, and optimizes a trajectory interpolant under this metric. PBDMP is
a trajectory-based diffusion motion planner that additionally takes the scene layout
as input to enable generalization across environments. In contrast, \textbf{\name} is trained
only on state observations and constructs trajectories at inference time directly from
the learned local geometry of the state space.

\section{Robustness to Score-Model Retraining}
\label{sec:seed_robustness}

Although trajectory construction is deterministic given a trained generative model,
the learned state-space geometry may vary with model initialization and optimization.
We therefore independently train five score models and evaluate \textbf{\name} with each model
on the three DualKUKA settings. The results are reported in
Table~\ref{tab:seed_robustness}.

\begin{table}[t]
  \caption{
  Robustness across five independently trained score models on DualKUKA.
  We report mean $\pm$ standard deviation of success rate (\%).
  }
  \label{tab:seed_robustness}
  \begin{center}
    \begin{small}
      \begin{sc}
        \begin{tabular}{lccc}
          \toprule
          Method
          & \makecell{$n=7$\\$s=0.14$}
          & \makecell{$n=5$\\$s=0.14$}
          & \makecell{$n=5$\\$s=0.22$} \\
          \midrule
          \textbf{\name}
          & $56.90 \pm 2.22$
          & $50.70 \pm 3.25$
          & $51.30 \pm 1.99$ \\
          \bottomrule
        \end{tabular}
      \end{sc}
    \end{small}
  \end{center}
\end{table}

The relatively small variation across independently trained models indicates that
\textbf{\name}'s planning performance is not strongly dependent on a particular score-model
initialization.


\newpage
\section*{NeurIPS Paper Checklist}

The checklist is designed to encourage best practices for responsible machine learning research, addressing issues of reproducibility, transparency, research ethics, and societal impact. Do not remove the checklist: {\bf The papers not including the checklist will be desk rejected.} The checklist should follow the references and follow the (optional) supplemental material.  The checklist does NOT count towards the page
limit. 

Please read the checklist guidelines carefully for information on how to answer these questions. For each question in the checklist:
\begin{itemize}
    \item You should answer \answerYes{}, \answerNo{}, or \answerNA{}.
    \item \answerNA{} means either that the question is Not Applicable for that particular paper or the relevant information is Not Available.
    \item Please provide a short (1--2 sentence) justification right after your answer (even for \answerNA). 
\end{itemize}

{\bf The checklist answers are an integral part of your paper submission.} They are visible to the reviewers, area chairs, senior area chairs, and ethics reviewers. You will also be asked to include it (after eventual revisions) with the final version of your paper, and its final version will be published with the paper.

The reviewers of your paper will be asked to use the checklist as one of the factors in their evaluation. While \answerYes{} is generally preferable to \answerNo{}, it is perfectly acceptable to answer \answerNo{} provided a proper justification is given (e.g., error bars are not reported because it would be too computationally expensive'' or ``we were unable to find the license for the dataset we used''). In general, answering \answerNo{} or \answerNA{} is not grounds for rejection. While the questions are phrased in a binary way, we acknowledge that the true answer is often more nuanced, so please just use your best judgment and write a justification to elaborate. All supporting evidence can appear either in the main paper or the supplemental material, provided in appendix. If you answer \answerYes{} to a question, in the justification please point to the section(s) where related material for the question can be found.

IMPORTANT, please:
\begin{itemize}
    \item {\bf Delete this instruction block, but keep the section heading ``NeurIPS Paper Checklist"},
    \item  {\bf Keep the checklist subsection headings, questions/answers and guidelines below.}
    \item {\bf Do not modify the questions and only use the provided macros for your answers}.
\end{itemize}


\begin{enumerate}

\item {\bf Claims}
    \item[] Question: Do the main claims made in the abstract and introduction accurately reflect the paper's contributions and scope?
    \item[] Answer: \answerYes{} 
    \item[] Justification: The contributions and scope are well covered in the following sections of Introduction.

\item {\bf Limitations}
    \item[] Question: Does the paper discuss the limitations of the work performed by the authors?
    \item[] Answer: \answerYes{} 
    \item[] Justification: Limitation is discussed in Conclusion.

\item {\bf Theory assumptions and proofs}
    \item[] Question: For each theoretical result, does the paper provide the full set of assumptions and a complete (and correct) proof?
    \item[] Answer: \answerNA{} 
    \item[] Justification: The paper contains no theoretical results but only uses theoretical hypotheses to guide the development of the methodology.

    \item {\bf Experimental result reproducibility}
    \item[] Question: Does the paper fully disclose all the information needed to reproduce the main experimental results of the paper to the extent that it affects the main claims and/or conclusions of the paper (regardless of whether the code and data are provided or not)?
    \item[] Answer: \answerYes{} 
    \item[] Justification: The hyperparameters are provided in the Appendix and source code will be made public upon acceptance.

\item {\bf Open access to data and code}
    \item[] Question: Does the paper provide open access to the data and code, with sufficient instructions to faithfully reproduce the main experimental results, as described in supplemental material?
    \item[] Answer: \answerNo{} 
    \item[] Justification: We will release the code upon acceptance.

\item {\bf Experimental setting/details}
    \item[] Question: Does the paper specify all the training and test details (e.g., data splits, hyperparameters, how they were chosen, type of optimizer) necessary to understand the results?
    \item[] Answer: \answerYes{} 
    \item[] Justification: Experimental details are provided in the Appendix

\item {\bf Experiment statistical significance}
    \item[] Question: Does the paper report error bars suitably and correctly defined or other appropriate information about the statistical significance of the experiments?
    \item[] Answer: \answerYes{} 
    \item[] Justification:We report mean and standard deviation across five independently trained score models for the DualKUKA experiments in Appendix~\ref{sec:seed_robustness}.

\item {\bf Experiments compute resources}
    \item[] Question: For each experiment, does the paper provide sufficient information on the computer resources (type of compute workers, memory, time of execution) needed to reproduce the experiments?
    \item[] Answer: \answerYes{} 
    \item[] Justification: The full details of compute resources is provided in the Appendix.
    
\item {\bf Code of ethics}
    \item[] Question: Does the research conducted in the paper conform, in every respect, with the NeurIPS Code of Ethics \url{https://neurips.cc/public/EthicsGuidelines}?
    \item[] Answer: \answerYes{} 
    \item[] Justification: Code of Ethics is fully understood and conformed.

\item {\bf Broader impacts}
    \item[] Question: Does the paper discuss both potential positive societal impacts and negative societal impacts of the work performed?
    \item[] Answer: \answerNA{} 
    \item[] Justification: The work includes no social aspects but focuses on core deep learning problems.
    
\item {\bf Safeguards}
    \item[] Question: Does the paper describe safeguards that have been put in place for responsible release of data or models that have a high risk for misuse (e.g., pre-trained language models, image generators, or scraped datasets)?
    \item[] Answer: \answerNA{} 
    \item[] Justification: The paper poses no such risks.

\item {\bf Licenses for existing assets}
    \item[] Question: Are the creators or original owners of assets (e.g., code, data, models), used in the paper, properly credited and are the license and terms of use explicitly mentioned and properly respected?
    \item[] Answer: \answerYes{} 
    \item[] Justification: All the code and data used are properly cited in the paper.

\item {\bf New assets}
    \item[] Question: Are new assets introduced in the paper well documented and is the documentation provided alongside the assets?
    \item[] Answer: \answerNA{} 
    \item[] Justification: Code will be made public upon acceptance. Other than the code, the paper contains no new assets.

\item {\bf Crowdsourcing and research with human subjects}
    \item[] Question: For crowdsourcing experiments and research with human subjects, does the paper include the full text of instructions given to participants and screenshots, if applicable, as well as details about compensation (if any)? 
    \item[] Answer: \answerNA{} 
    \item[] Justification: No human is included in the experiments.

\item {\bf Institutional review board (IRB) approvals or equivalent for research with human subjects}
    \item[] Question: Does the paper describe potential risks incurred by study participants, whether such risks were disclosed to the subjects, and whether Institutional Review Board (IRB) approvals (or an equivalent approval/review based on the requirements of your country or institution) were obtained?
    \item[] Answer: \answerNA{} 
    \item[] Justification: No human subjects or crowdsourcing is included in the paper.

\item {\bf Declaration of LLM usage}
    \item[] Question: Does the paper describe the usage of LLMs if it is an important, original, or non-standard component of the core methods in this research? Note that if the LLM is used only for writing, editing, or formatting purposes and does \emph{not} impact the core methodology, scientific rigor, or originality of the research, declaration is not required.
    \item[] Answer: \answerNA{} 
    \item[] Justification: The research does not involve LLMs as any important, original, or non-standard components.

\end{enumerate}

\end{document}